# A Blended Likelihood Approach for Achieving Fairness Using Naïve Bayes

John Arthur Junior[1], Abdul Lateef Yussif[1], Maame G. Asante-Mensah[1], Charles R. Haruna[1], Sandro Amofa[1], Elliot Attipoe[1].

1 Department of Computer Science and Information Technology, University of Cape Coast, Cape Coast, Ghana.

Address all correspondence to:

**Maame G. Asante-Mensah**

gasante-mensah@ucc.edu.gh

## Abstract

Concerns about algorithmic bias and fairness have increased as artificial intelligence has been incorporated into high-stakes decision-making. Traditional Naïve Bayes classifiers, while efficient and interpretable, lack fairness-awareness mechanisms and perpetuate historical biases in sensitive domains such as hiring, credit scoring, and criminal justice. This study develops a fairness-aware extension of the Naïve Bayes classifier that mitigates bias while maintaining computational efficiency. We propose the Bias Mitigating Naïve Bayes (BMNB) classifier, integrating in-processing and post-processing interventions. The in-processing stage employs a blended likelihood approach combining group-specific and pooled likelihood estimates through a tunable blending parameter ($\alpha$) to balance fairness and accuracy. The post-processing stage applies output calibration with adaptive thresholding to fine-tune group-specific decision boundaries. Experimental results indicate that BMNB attains Disparate Impact (DI) values of 1.000, 1.171, and 0.997 and Equal Opportunity Difference (EOD) values of −0.217, −0.226, and −0.053 on the Adult, ProPublica, and Framingham datasets, respectively, while maintaining computational efficiency. Ablation studies confirm that the combination of blended likelihood and adaptive thresholding yields superior performance compared to either technique in isolation.

## Keywords

Adaptive Thresholding, Bias Mitigation, Bias Mitigating Naïve Bayes (BMNB), Blended Likelihood Estimation (BLE), Naïve Bayes (NB), Machine Learning (ML)

## 1. Introduction

ML systems are increasingly used in socially critical domains such as hiring, credit scoring, healthcare, and criminal justice, with issues of bias and fairness having emerged. Systems expected to make objective decisions have been shown to exhibit biased behaviour, potentially aggravating existing inequities embedded in historical data, leading to discriminatory outcomes [1], [2], [3]. High-profile cases demonstrating such disparities include racial bias in the COMPAS recidivism prediction tool's algorithm which falsely predicts future criminality among African Americans at a rate twice as high as for white individuals [1], [4], [5], [6]. Amazon identified discrimination in its AI hiring system, specifically against female candidates, particularly in the context of software development and technical positions. One possible reason for the gender bias in AI systems is the overrepresentation of data from male software developers in the training data [7]. Existing research on fairness has explored a wide range of strategies across the ML pipeline including pre-processing, in-processing and post-processing techniques [1], [8], [9]. As complex models such as deep neural networks and logistic regression have received considerable attention for fairness interventions, traditional algorithms like NB remain relatively underexplored despite their continued relevance in real-world applications due to their simplicity and efficiency [10], [11]. NB is a simple yet effective probabilistic classifier widely used for high-dimensional data. However, standard NB assumes feature independence and lacks mechanisms to ensure fairness, making it vulnerable to disparate impact when trained on historic datasets [12]. Previous fairness studies such as N-Naïve Bayes and Fair-CMNB, has largely applied single-stage interventions either preprocessing or post-processing [13], [14], rather than designing systematic unified multi-stage adaptations of the ML pipeline. This represents a significant gap in the literature where no unified fairness-aware model has been proposed that seamlessly combines in-processing and post-processing techniques tailored to NB. This paper introduces the Bias Mitigating Naïve Bayes (BMNB), a unified model that addresses bias in both the training and decision stages of NB. The approach integrates blended likelihood estimation (in-processing) with adaptive thresholding (post-processing) to mitigate disparities while retaining BMNB's interpretability and efficiency. Guided by the methodological gap identified above, the study is driven by the following research questions: To what extent does the proposed BMNB classifier reduce algorithmic bias as measured through DI, EOD, EMOD, Bias Index and Fairness Score and compared to baseline NB across multiple benchmark datasets? How does varying the blending coefficient α in the BMNB model influence the fairness-accuracy trade-off, and what alpha (α) range provides the most balanced performance across different dataset? What are the relative contributions of the proposed techniques to the overall fairness improvement achieved by BMNB? The motivation for BMNB is further supported by empirical findings conducted on the Adult, ProPublica COMPAS and Framingham dataset showed significantly reduce group disparities while maintaining competitive performance. BMNB achieved DI values 1.000, 1.171 and 0.997 respectfully, demonstrating meaningful bias reduction across datasets.

## 2. Proposed BMNB Model

Figure 1 illustrates the systematic approach used in this study to address the algorithmic bias, highlighting the workflow from initial data acquisition and pre-processing through to model development and evaluation and finally a fair prediction.

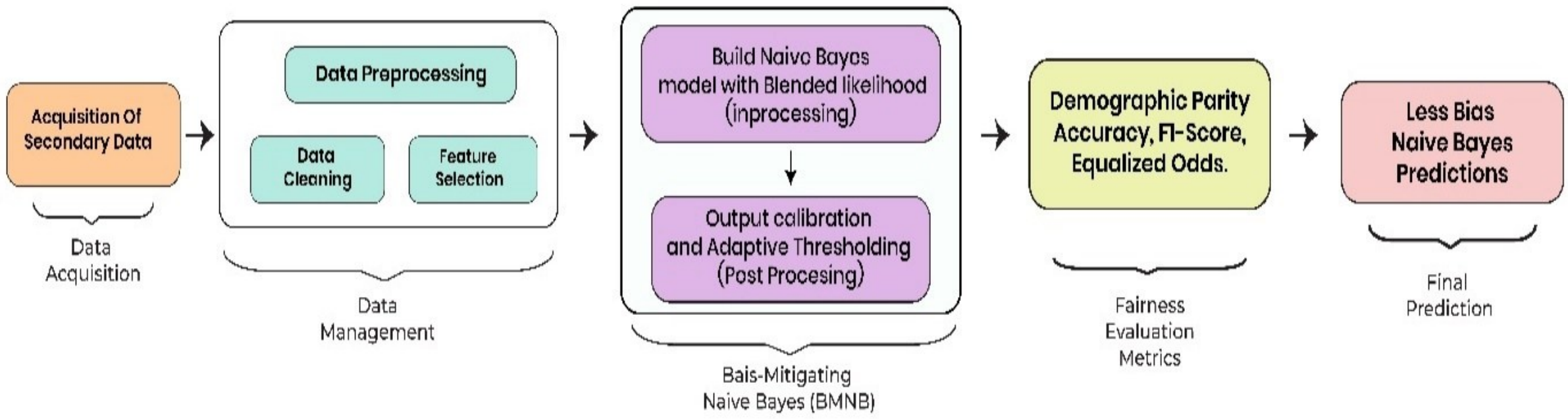

**Figure 1. Systematic Approach**

The model pipeline comprised three main components:

- Group-specific NB classifiers at the in-processing stage.
- Blended likelihood computation using a tunable blending coefficient α ∈ [0.00,0.25,0.50,0.75,1.00] to combine global and group-specific likelihood.
- Adaptive threshold calibration in the post-processing stage was applied separately for each sensitive group.

The α parameter determines the trade-off between global and group-specific patterns, while the adaptive threshold adjusts decision boundaries to align with the statistical characteristics of each group.

**2.1 Proposed Techniques Applied**

The BMNB model integrates two complementary fairness-enhancing techniques: Blended likelihood estimation (in-processing) [13], [14] and Adaptive threshold calibration (post-processing) [15]. In the in-processing stage, separate NB models are trained for each subgroup of sensitive attributes, and their predicted probabilities are blended with those from a global model using a tunable blending coefficient (α). This enables a control balance between group-specific and overall patterns during learning. Following this, the post-processing stage, group-specific decision thresholds are calibrated to align with fairness criteria such as DP or EO, thereby reducing residual disparities after training. BMNB provides a unified and deployable classifier for fairness-aware classification while retaining the transparency and efficiency of the traditional NB algorithm as illustrated in Figure 2.

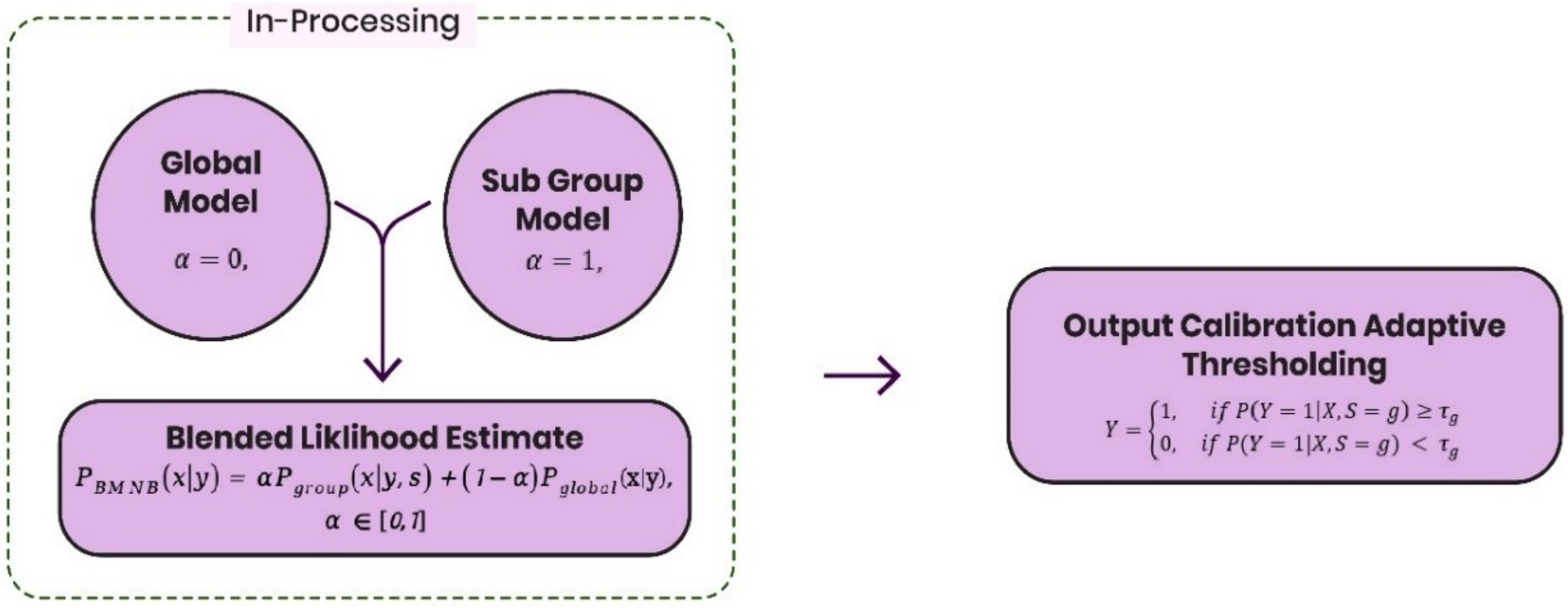

**Figure 2. Proposed BMNB Technique**

### 2.1.1 Blended Likelihood Estimation (BLE)

The BLE component of BMNB modifies the standard NB probability estimation process to incorporate both group-specific and global information. In the conventional case in NB, the posterior probability of a class (y) given a feature vector (x) is computed as [16]:

$$P(y|x) \propto P(y) \prod_i P(x_i|y) \tag{1}$$

In BMNB, this likelihood is extended by introducing a blending parameter (α) that linearly combines the group-specific and global likelihoods as:

$$P_{BMNB}(x|y) = \alpha P_{group}(x|y,s) + (1-\alpha) P_{global}(x|y), \quad \alpha \in [0,1] \tag{2}$$

Where: $P_{group}(x|y,s)$: likelihood estimates from a subgroup.

$P_{global}(x|y)$: likelihood estimate across all data samples.

$\alpha$: blending parameter $0 \leq \alpha \leq 1$.

When $\alpha = 0$, the model behaves like a traditional global NB, when $\alpha = 1$, it fully relies on subgroup-specific distributions. Intermediate values of α create a bias-variance trade-off, allowing the model to balance fairness and overall predictive performance. In the training phase, the BMNB algorithm computes class-conditional probabilities separately for each sensitive subgroup and for the global dataset. During prediction, it blends these probabilities according to the optimal α value determined through cross-validation or fairness-accuracy evaluation.

---

***Algorithm for Naïve Bayes with Blended Likelihood***

---

***Input:*** *training features* $X_{train}$, *training labels,* $Y_{train}$, *training group attributes* $S_{train}$ *test features* $X_{test}$ *test group attributes* $S_{test}$, *blending coefficient* $\alpha \in [0,1]$ *(or α_grid for selection), smoothing parameter* $\varepsilon > 0$

***Target:*** *Predicted labels* $Y_{pred}$

*1.* $G = \{s_i \mid (x_i, s_i) \in S_{train}\}$

2. $D_{global} = (X_{train}, Y_{train}) = \{(x_i, y_i)\}_{i=1}^{N_{train}}$

*3.* $M_{global} = Train_{NB}(D_{global}; \varepsilon)$

4. $\forall g \in G$:

$$D_g = \{(x_i, y_i) \mid s_i = g, (x_i, s_i, y_i) \in S_{train}\}$$

$$M_g = Train_{NB}(D_g; \varepsilon)$$

5. $If\ \alpha\ not\ provided$:

$$\alpha^* = \underset{\alpha \in \alpha_{grid}}{argmax}\ J(Acc_{CV}(\alpha), Fair_{CV}(\alpha))$$

$$J(Acc, Fair) = \lambda \cdot Acc + (1-\lambda) \cdot Fair, \qquad \lambda \in [0,1]$$

$\alpha \leftarrow \alpha^*$

6. $Y_{pred} = \emptyset$

7. $\forall x_i \in X_{test}$:

$g = s_i \in S_{test}$

$P_{group} = \begin{cases} M_{global}.predict_{proba(x_i)}, & if\ g \notin G\ or\ M\ lower\ suport \\ M_g.predict_{proba(x_i)}, & otherwise \end{cases}$

$P_{group} = M_{global}.predict_{proba(x_i)}$

$P_{BMNB} = aP_{group} + (1-a)P_{global}$

$\hat{y}_i = \begin{cases} argmax_y\ P_{BMNB}[y], & if\ T_g\ not\ used \\ 1, & if\ P_{BMNB}[y_i] > T_g \end{cases}$

$Y_{pred} \longleftarrow Y_{pred} \cup \{\hat{y}_i\}$

*8. Return* $Y_{pred}$

### 2.1.2 Output Calibration and Adaptive Thresholding

Unlike conventional methods that use uniform thresholds for all groups, this approach adjusts thresholds separately for each sensitive group so that fairness criteria such as DP or EO are satisfied [24]. Group-specific thresholds are computed based on performance metrics such as TPR and adapted to equalize the chosen fairness metrics across groups [17]. This method operates without retraining the model making it suitable for modifying the training process. Generally, for a standard binary classifier, a decision threshold ($\tau$) is applied as:

$$Y = \begin{cases} 1, & if\ P(Y=1|X) \geq \tau \\ 0, & if\ P(Y=1|X) < \tau \end{cases} \tag{3}$$

To address disparities, we introduce group-specific thresholds ($\tau_g$) for each sensitive group g:

$$Y = \begin{cases} 1, & if\ P(Y=1|X,S=g) \geq \tau_g \\ 0, & if\ P(Y=1|X,S=g) < \tau_g \end{cases} \tag{4}$$

Applying a fairness metric, EOD, measured in terms of TPR for each group:

$$TPR_g = P\left(\hat{Y} = 1 \middle| Y = 1, S = g\right) \tag{5}$$

$$EOD = \left|TPR_{g1} - TPR_{g2}\right| \tag{6}$$

We apply a score-based post-processing step in the spirit of [24] where group-specific thresholds are derived from each group's predicted score distribution. By selecting quantile-based thresholds that match the overall positive rate, the method produces monotone group-conditional decision rules that adjust only the final classification stage.

This adaptive thresholding approach enables fairness-aware decision-making, allowing practitioners to fine-tune group outcomes and reduce residual bias in ML pipelines.

## 2.2 Summary of the Proposed Fairness Pipeline

The fairness enhancement pipeline for bias mitigation illustrated in Figure 3 presents the conceptual workflow of the proposed design. It depicts a sequential and interconnected process in which each stage contributes to bias reduction. Preprocessing focuses on data cleaning, transformation, and rebalancing to minimize inherent biases in the dataset. The in-processing stage incorporates the blended likelihood approach, where group-specific and global probability estimates are combined using a tunable parameter (α), enabling a controlled trade-off between fairness and accuracy during model training. Subsequently, the post-processing stage applies adaptive thresholding, whereby group-specific decision thresholds are calibrated to satisfy fairness criteria such as demographic parity or equal opportunity. This step mitigates residual bias in model outputs without requiring retraining. The color-coded structure of the pipeline distinguishes the different categories of fairness interventions, emphasizing their complementary roles within a unified framework. The final output of the pipeline is an unbiased prediction, reflecting decisions that have been systematically adjusted to reduce disparities across sensitive groups. Overall, the proposed framework demonstrates that integrating fairness mechanisms at multiple stages of the pipeline provides a more effective and practical approach to bias mitigation than single-stage methods.

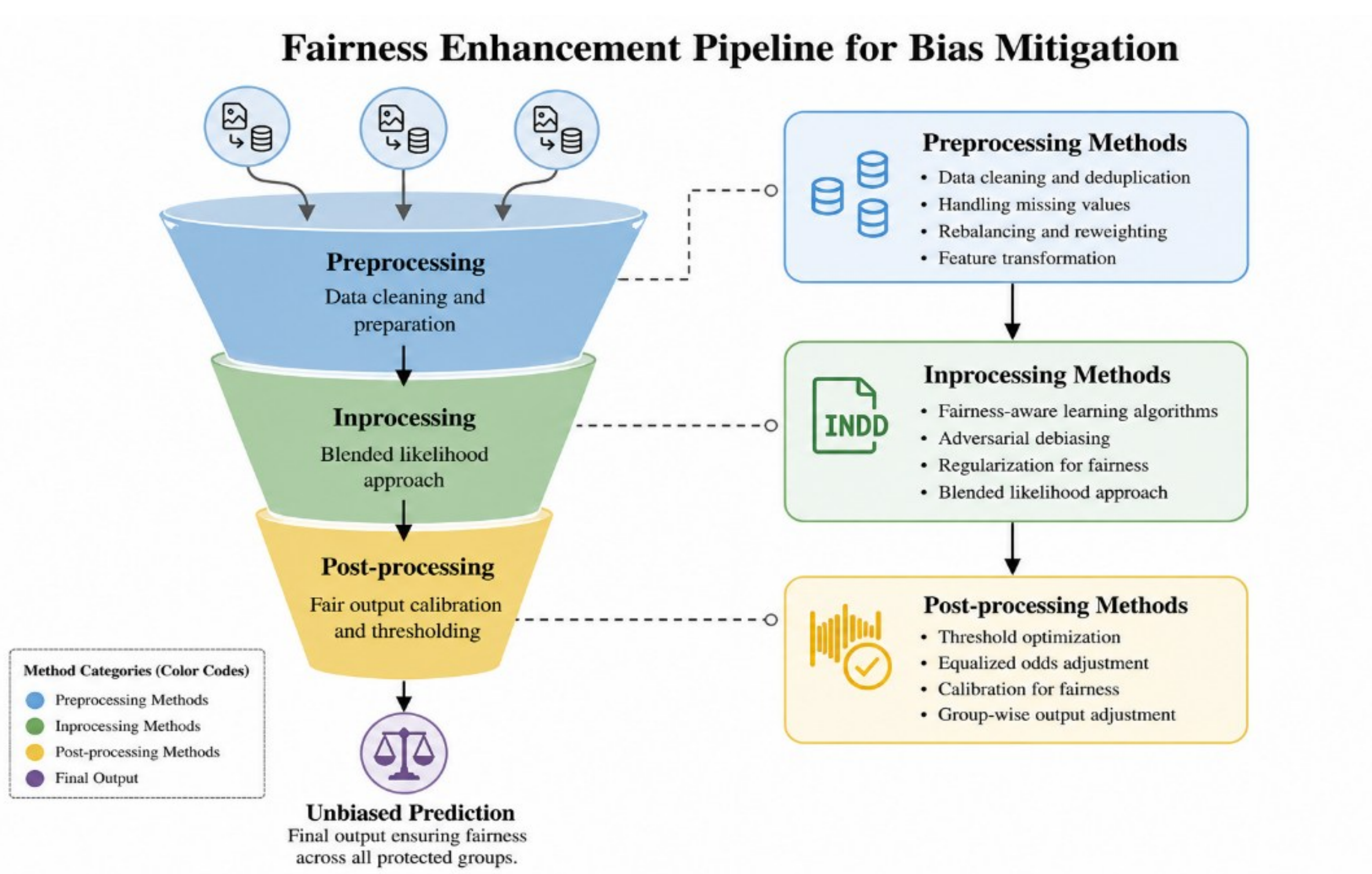

**Figure 3 A proposed fairness pipeline.**

## 3. Experiments and Results

The BMNB classifier was implemented and tested on the Adult, ProPublica COMPAS, and Framingham datasets. The Adult dataset from the UCI Machine Learning Repository predicts income level, with gender serving as the sensitive attribute. The ProPublica COMPAS dataset is used for predicting recidivism risk and considers race as the sensitive attribute. The Framingham Heart Study dataset predicts the 10-year risk of developing cardiovascular disease, with gender as the sensitive attribute. Table 1 provides a summary description of the experimental dataset. For each dataset, an 80-20 stratified train-test split was applied to maintain the distribution of both labels and sensitive groups in each partition. Data pre-processing involved scaling all numerical features to the (0,1) range to ensure uniform feature influence, applying SMOTEEN to address zero-frequency issues, class imbalance and prevent numerical underflow if any, one-hot encoding categorical variables to prevent the imposition of ordinal relationships, and handling missing values according to dataset-specific requirements. A GaussianNB classifier with fixed class priors was used. To ensure a fair comparison, the same pre-processed datasets and data partitions were used to train the baseline NB. A full replication package including full preprocessing script, training code, evaluation pipeline and experiment configurations is publicly available at Github.

| Dataset | Task | Sensitive Attribute | Class Imbalance | Key Features |
|---|---|---|---|---|
| Adult Income | Income > 50K prediction | Gender | 24% vs 76% | Age, Education, Occupation |
| Framingham Heart Study | 10-year CHD risk prediction | Gender | Moderate | Blood pressure, cholesterol |
| ProPublica COMPAS | 2-year recidivism prediction | Race | Severe | Prior convictions, age |

**Table 1. Summary Description of Experimental Datasets**

### 3.1 Evaluation Metrics of the Proposed BMNB Method

To assess the effectiveness of the proposed BMNB in mitigating biased predictions and ensuring fairness, a set of carefully selected evaluation metrics has been chosen. Fairness metrics provide a comprehensive understanding of the model's ability to balance fairness and predictive accuracy, particularly in addressing disparities affecting minority groups [18]. The work by [19] highlights the need for tailored fairness measures that account for specific regulatory and contextual requirements, as no single metric can universally address all fairness concerns [25].

#### 3.1.1 Disparate Impact (DI)

DI is a key fairness metric that evaluates whether a machine learning model disproportionately favours or disadvantages certain demographic groups [20]. It is particularly relevant in high-stakes decision-making systems, such as hiring, lending, and criminal justice, where biased models can reinforce systemic inequalities [21].

Mathematically, DI is defined as the ratio of positive prediction rates (selection rates) between two demographic groups:

$$DI = \frac{P(Y=1|S=s_1)}{P(Y=1|S=s_2)} \tag{7}$$

Where: $P(Y=1|S=s)$ represents a positive prediction rate.

$s_1$, $s_2$ are distinct values of the sensitive attribute.

If $DI = 1$, the model treats both groups equally,

If $DI < 1$, one group has fewer positive predictions, it indicates potential discrimination.

### 3.1.2 Equal Mis-Opportunity Difference (EMOD)

This measures whether a machine learning model makes false positive errors disproportionately across different sensitive groups (e.g., racial or gender groups). It specifically compares the FPR between two groups [22].

$$EMOD = FPR_{unprivileged} - FPR_{privileged} \tag{8}$$

Where:

$$FPR = \frac{False\ Positives}{False\ Negatives + True\ Negatives} \tag{9}$$

Unprivileged group: the historically disadvantaged group (e.g., African-Americans, women)

Privileged group: a group that has more social or systemic advantages

$EMOD = 0$, perfect fairness in false positives across groups.

$EMOD > 0$, higher FPR for the unprivileged group: an unfair disadvantage.

$EMOD < 0$, lower FPR for the unprivileged group.

The closer to 0, the fairer the model regarding false positives.

### 3.1.3 Equal Opportunity Difference (EOD)

EOD is a widely adopted group fairness metric that measures disparities in TPR's between sensitive subgroups [23]. It evaluates whether individuals who belong to different demographic groups (e.g., by race or gender) but share the same actual outcome are equally likely to receive a positive prediction from the model.
Mathematically, EOD is defined as the difference in true positive rates (also called recall) between the unprivileged and privileged groups.

$$EOD = TPR_{unprivileged} - TPR_{privileged} \tag{10}$$

Where:

$$TRP = \frac{True\ Positives}{True\ Positives + False\ Negatives} \tag{11}$$

The unprivileged group typically refers to a historically marginalized or disadvantaged group.

The privileged group refers to a group that has traditionally benefited from systemic advantage.

$EOD = 0$, both groups have equal chance of receiving true positive predictions.

$EOD > 0$, privileged group has higher true positive (unfair advantage).

$EOD < 0$, unprivileged group has higher true positive (reverse bias).

### 3.1.4 Bias Index (BI)

BI and Fairness Score are novel proposals for measuring fairness in AI systems proposed by [21]. The BI is introduced to quantify the degree of bias for each protected attribute within a dataset or AI model. It helps in identifying how much bias exists for specific sensitive characteristics like gender, race, or marital status. It is calculated based on the deviation of individual fairness metrics (like SPD or DI) from their ideal values for each protected attribute. The formula for BI for a protected attribute (i) is given as:

$$BI_i = \sqrt{\frac{1}{n}\sum_{j=1}^{n}\left(M_j - M_j^i\right)^2} \tag{12}$$

For a fair system, the BI for each protected attribute should ideally be zero. A value between 0 and 0.1 indicates that the system is unbiased for that attribute.

### 3.1.5 Fairness Score (FS)

The FS is proposed as a single, overall measure of fairness for the entire AI system, considering all protected attributes collectively. It provides a standardized linear scale for assessing fairness across different AI systems, enabling comparison. The FS is derived from the Bias Indexes of all protected attributes. It is defined for the AI system as:

$$FS = 1 - \sqrt{\frac{1}{n}\sum_{j=1}^{n}\left(M_j - M_j^i\right)^2} \tag{13}$$

For a perfectly fair system, the FS should be one. A fairness score between 0.9 and 1.0 indicates a fair AI system.

## 3.2 Experiment 1: The BMNB Model on the Benchmark Datasets

This experiment evaluated the performance of the proposed BMNB classifier across three benchmark datasets as described in Table 1. The objective is to determine whether BMNB can reduce group disparities. We report classification metrics (accuracy, precision, recall, F1-score) and fairness indicators BI and FS. The results summarized in Table 2, shows that BMNB achieved balance predictive performance across all task while substantially improving fairness. BI values were low (Adult: 0.091; ProPublica: 0.188; Framingham: 0.021) corresponding to high FS (0.909, 0.812, 0.979) respectively. These findings indicate that BMNB effectively reduced group disparities without compromising model's overall ability to predict accurately.

| | Adult Dataset | | ProPublica Dataset | | Framingham Dataset | |
|---|---|---|---|---|---|---|
| **Target** | **<=50K** | **>50K** | **No Recidivism** | **Recidivism** | **No Risk** | **At Risk** |
| Precision | 0.61 | 0.87 | 0.82 | 0.82 | 0.51 | 0.85 |
| Recall | 0.87 | 0.61 | 0.80 | 0.84 | 0.85 | 0.51 |
| F1 Score | 0.71 | 0.71 | 0.81 | 0.83 | 0.64 | 0.63 |
| Support | 1798 | 2574 | 164 | 180 | 342 | 576 |
| Accuracy | 0.71 | | 0.82 | | 0.64 | |
| Bais Index (BI) | 0.091 | | 0.188 | | 0.021 | |
| Fairness Score (FS) | 0.909 | | 0.812 | | 0.979 | |

**Table 2. Results from Proposed BMNB on Benchmark Datasets**

### 3.3 Experiment 2: Varying Blending Parameter Alpha ($\alpha$ )

The effect of varying the blending coefficient (α) proposed in equation 2 was systematically examined to understand how it shapes the fairness-accuracy dynamics of the BMNB model. The parameter α was introduced as a fairness-aware control to interpolate between global likelihoods (α=0.0) and group-specific likelihoods (α =1.0), thereby influencing the degree of fairness in the classification decision-making process. Table 3 provides a summary of the effect of alpha. Experimental results, as depicted in Figure 4, illustrate how α modifies the fairness-accuracy trade-off across all three datasets. Specifically, increasing α favours overall predictive accuracy but may reduce fairness across sensitive groups, while decreasing α promotes fairness at the potential expense of accuracy. This behaviour validates the intended role of α in the in-processing phase of the BMNB model and provides empirical support for the model's flexibility in navigating between competing objectives of performance and equity. From Figure 3(a), ProPublica and Framingham decreased in accuracy (from 0.820 to 0.770) and (0.64 to 0.620) respectively with α ranging from (0.80 – 1.00) which further asserts the claims.

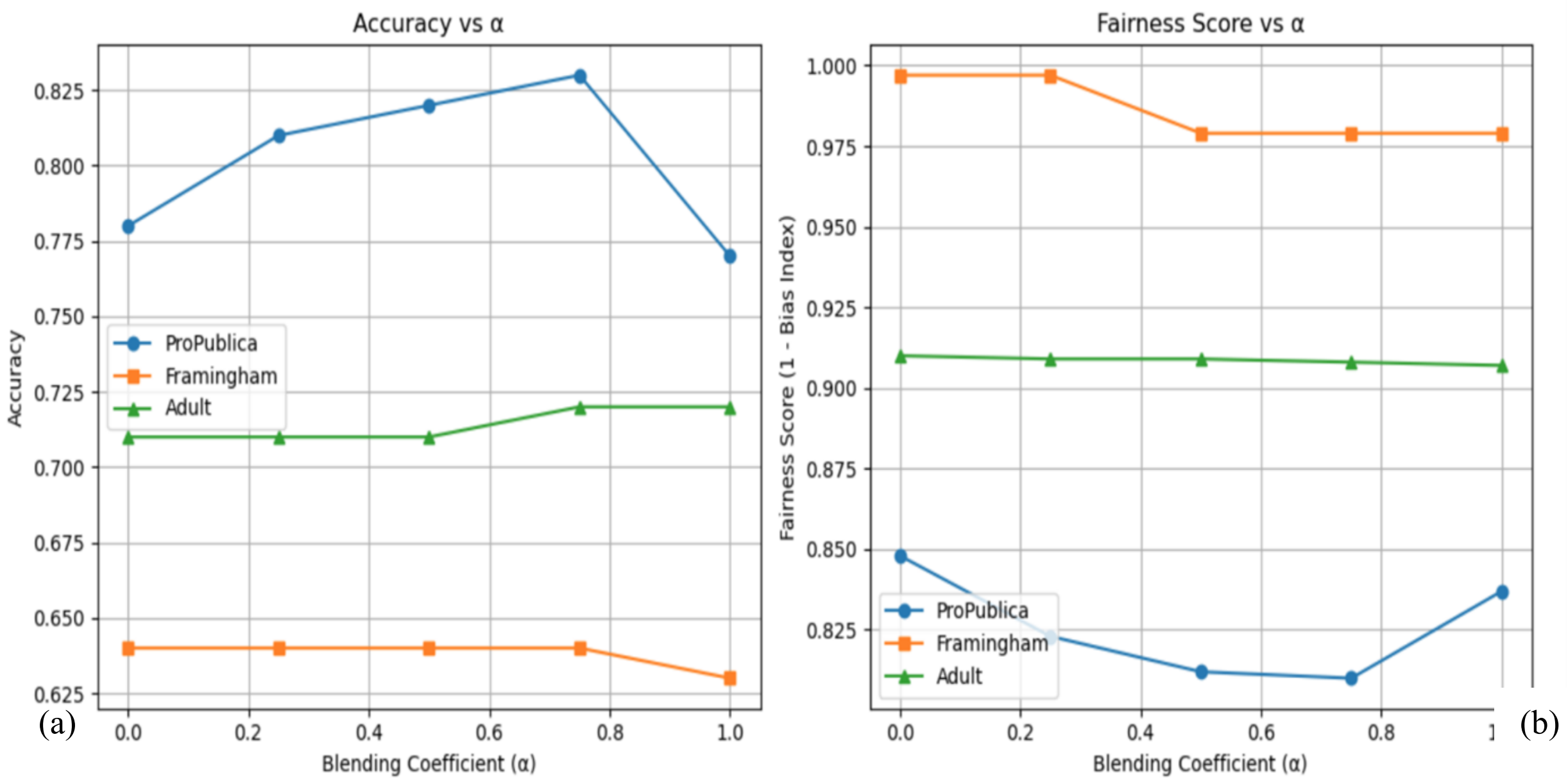

**Figure 4. Accuracy Vs Alpha Parameters and Fairness Score Vs Alpha Parameters**

| Alpha ($\alpha$) | General Description |
|---|---|
| Greater than ($> 0.5$) | an increased influence of one group model, potentially amplifying existing bias. |
| Equal to ($= 0.5$) | provides balanced treatment of both groups, ensuring equal contribution from group-specific models. |
| Less than ($< 0.5$) | much emphasis on the other groups characteristics, sometimes overcompensating for fairness. |

**Table 3. Overall Summary of Effect of Alpha**

### 3.4 Evaluation of BMNB using Bias Index and Fairness Score

The comparative evaluation of the proposed BMNB model and the fairness framework of [21] on the adult dataset (gender bias) reveals notable differences in fairness performance across multiple metrics. From the results, BMNB achieves SPD = 0.000 and DI = 1.000, both indicating perfect parity between male and female subgroups. In contrast, [21] reports SPD = -0.1945 and DI = 0.3598, showing measurable disparities in group outcomes. These results demonstrate that BMNB substantially reduces group-level disparities compared to [21] approach. On error-based metrics, BMNB records EOD = -0.217 and EMOD = -0.148, which although non-zero, remain within acceptable fairness thresholds. Conversely, [21] results yield EOD = 0.1257 and EMOD = 0.0958, showing lower error disparities relative to BMNB. This indicates that while BMNB excels at ensuring group-level fairness in outcomes (SPD, DI), [21] approach performs slightly better in balancing error rates between subgroups (EOD, EMOD). A consolidated fairness measure through BI and FS highlights the trade-off between the two approaches. BMNB attains a BI of 0.0913 and FS of 0.9088, demonstrating an overall fairer model compared to Agarwal's BI of 0.2641 and FS of 0.7360. This underscores BMNB's advantage in reducing residual bias and achieving higher aggregate fairness. The comparison shows that BMNB provides stronger group-level fairness guarantees (parity in outcomes), whereas [21] framework balances error-rate fairness more effectively. However, BMNB outperforms overall when considering the aggregated BI and FS, suggesting that it offers a more robust mitigation strategy for gender bias in the adult dataset, as indicated in Table 3.

| Adult Dataset | | | |
|---|---|---|---|
| **Fairness Metrics** | **BMNB** | **Fairness Certification [21]** | Optimal Value |
| SPD | 0.000 | -0.1945 | 0 |
| DI | 1.000 | 0.3598 | 1 |
| EOD | -0.217 | 0.1257 | 0 |
| EMOD | -0.148 | 0.0958 | 0 |
| BI | **0.0913** | 0.2641 | 0 |
| FS | **0.9088** | 0.7360 | 1 |

**Table 4. Comparative Analysis of BMNB and Fairness Certification [21] on Fairness Metrics**

These results indicate that the BMNB model achieves a high degree of fairness, with minimal deviation from fairness ideals across all evaluated dimensions. This provides empirical support for the model's ability to balance fairness with predictive performance, aligning with the objective of mitigating algorithmic bias across sensitive subgroups.

### 3.5 Fairness-Accuracy Trade-off

The experimental results revealed a consistent trade-off between fairness and accuracy when applying the BMNB model. It is observed from Figure (4b) that higher α values (approaching 1.0) maximize fairness adjustments, but sometimes at the cost of overall accuracy. Lower α values (approaching 0.0) maintained higher accuracy but provided minimal fairness improvements. The optimal range of α, which balances fairness and accuracy, was found to range between 0.3 and 0.7, although the precise value varied depending on the dataset and sensitive attribute under consideration. These findings are consistent with prior fairness research, where scholars such as [10], [11] emphasize that algorithmic fairness and predictive utility are often in tension, and that eliminating disparities may not always be feasible without some performance sacrifice. The results further align with the broader discourse in fairness-aware machine learning, where the fairness-accuracy trade-off has been described as an unavoidable "cost of fairness" [12]. There is an inverse, non-linear, relationship between accuracy and fairness, a behaviour reported in existing literature. We introduce the proportional trade-off equation:

$$F * A = K \tag{14}$$

Where: F denotes Fairness, A denoted Accuracy and K is a dataset dependent constant.

### 3.6 Evaluation of BMNB and Baseline Naive Bayes Model.

Table 4 compares the baseline NB (without fairness intervention) with the proposed BMNB across the three datasets. The results show that while baseline NB achieved competitive accuracy and precision, it exhibited substantial disparities in fairness metrics. In contrast, BMNB consistently reduced bias, as reflected in lower BI values and higher FS across all datasets. For example, on the adult dataset, BMNB reduced BI from 0.224 to 0.091, improving the FS from 0.776 to 0.909, while also increasing recall and overall F1-score. On the ProPublica dataset, where baseline NB showed severe disparities (BI = 0.444, FS = 0.556), BMNB reduced the bias nearly by half (BI = 0.188, FS = 0.812). On the Framingham dataset, where fairness disparities were smaller, BMNB further improved fairness (BI = 0.021, FS = 0.979) without sacrificing accuracy. Overall, these results demonstrate that BMNB provides a better balance between predictive performance and fairness than the baseline NB, achieving values closer to the optimal fairness targets.

| | Adult Dataset | | ProPublica Dataset | | Framingham Dataset | |
|---|---|---|---|---|---|---|
| **Target** | **BL NB** | **BMNB** | **BL NB** | **BMNB** | **BL NB** | **BMNB** |
| Accuracy | 0.6699 | 0.7143 | 0.8895 | 0.8226 | 0.6307 | 0.6350 |
| Precision | 0.9414 | 0.8686 | 0.8776 | 0.8216 | 0.8434 | 0.8533 |
| Recall | 0.4685 | 0.6064 | 0.9166 | 0.8444 | 0.5052 | 0.5052 |
| F1 Score | 0.6256 | 0.7142 | 0.8967 | 0.8328 | 0.6319 | 0.6346 |
| EOD | -0.032 | -0.217 | 0.2159 | -0.226 | -0.095 | -0.053 |
| EMOD | 0.008 | -0.148 | -0.187 | -0.268 | -0.074 | -0.029 |
| SPD | 0.137 | 0.000 | 0.37 | 0.085 | -0.044 | -0.001 |
| DI | 1.717 | 1.000 | 2.0027 | 1.171 | 0.889 | 0.997 |
| BI | 0.224 | 0.091 | 0.4439 | 0.188 | 0.081 | 0.021 |
| FS | 0.776 | 0.909 | 0.5560 | 0.812 | 0.919 | 0.979 |

**Table 5. Comparison of baseline Naive Bayes and BMNB**

### 3.7 Ablation Study on the proposed BMNB across benchmark datasets

To isolate the contributions of each technique of the BMNB classifier, an ablation study was conducted on

- Blended Likelihood only (in-processing)
- Threshold Calibration only (post-processing)
- Full BMNB model combining all mechanisms

Each variant evaluated accuracy and fairness metrics (EOD, EMOD, DI, BI, FS) across the benchmark datasets. Across all datasets, the ablation shows in Tables 5,6 and 7 that threshold calibration is the primary driver of the fairness gains, consistently yielding lowest BI and highest FS. The blended likelihood alone provides moderate fairness improvement comparatively particularly on the highly biased ProPublica COMPAS dataset. The full BMNB model achieves the most balanced performance combining near optimal fairness with strong accuracy. Finding from the ablation study indicates that the integration of blended likelihood with group specific threshold calibration yields substantially greater bias mitigation than either component applied in isolation. This combined configuration consistently demonstrated improved fairness performance across metrics, indicating the complementary strengths of both mechanisms in BMNB.

| Method | Accuracy | EOD | EMOD | SPD | DI | BI | FS |
|---|---|---|---|---|---|---|---|
| Blended Only | 0.670 | -0.033 | 0.001 | 0.133 | 1.686 | 0.2130 | 0.7870 |
| Threshold Only | 0.716 | -0.225 | -0.144 | 0.000 | 1.000 | 0.0924 | 0.9076 |
| Full BMNB | 0.714 | -0.217 | -0.148 | 0.000 | 1.000 | 0.0914 | 0.9086 |

**Table 6: Ablation study on the adult dataset comparing blended likelihoods, threshold calibration, and the full BMNB method across accuracy and fairness metrics.**

| Method | Accuracy | EOD | EMOD | SPD | DI | BI | FS |
|---|---|---|---|---|---|---|---|
| Blended Only | 0.901 | 0.093 | -0.187 | 0.325 | 1.808 | 0.3532 | 0.6468 |
| Threshold Only | 0.770 | -0.345 | -0.268 | -0.013 | 0.974 | 0.1631 | 0.8369 |
| Full BMNB | 0.823 | -0.226 | -0.268 | 0.085 | 1.170 | 0.1872 | 0.8128 |

**Table 7: Ablation study on the ProPublica dataset comparing blended likelihoods, threshold calibration, and the full BMNB method across accuracy and fairness metrics.**

| Method | Accuracy | EOD | EMOD | SPD | DI | BI | FS |
|---|---|---|---|---|---|---|---|
| Blended Only | 0.617 | -0.113 | -0.073 | -0.056 | 0.847 | 0.0990 | 0.9010 |
| Threshold Only | 0.633 | -0.056 | -0.023 | -0.001 | 0.998 | 0.0204 | 0.9796 |
| Full BMNB | 0.635 | -0.053 | -0.030 | -0.001 | 0.998 | 0.0214 | 0.9786 |

**Table 8: Ablation study on the Framingham dataset comparing blended likelihoods, threshold calibration, and the full BMNB method across accuracy and fairness metrics.**

## 4. Acronyms and Abbreviations

Machine learning (ML), Naïve Bayes (NB), Blended Likelihood Estimation (BLE), Adaptive Thresholding, Bias Mitigation, Bias Mitigating Naïve Bayes (BMNB), Equal Opportunity Difference (EOD), Disparate Impact (DI), Demographic Parity (DP), Equal Opportunity (EO), Statistical Parity Difference (SPD), True Positive Rate (TPR), False Positive Rate (FPR).

## 5. Conclusion and Future Works

In this work, we propose a ready-to-implement classifier for practitioners seeking to improve fairness in NB classifiers. The BMNB classifier represents a novel approach to bias mitigation that combines the strengths of both in-processing and post-processing techniques, providing multiple layers of mitigation, unlike existing methods that typically focus on a single intervention point. The critical role of the blending coefficient (α) highlights the importance of tunability in fairness-aware algorithms. The BMNB classifier demonstrates that fairness improvements can be systematically introduced in a probabilistic model while still preserving acceptable levels of accuracy, provided that the blending parameter (α) is chosen judiciously. Our analysis reveals complex relationship between fairness and accuracy. We identified optimal α operating ranges as (0.25-0.75) approaching the pareto frontier of achievable fairness-accuracy combination. Future work should investigate integrating EOD-aware regularisation and post-processing calibration to mitigate dataset-dependent EOD trade-offs, and validate BMNB on additional diverse real-world cohorts. All research questions have been answered. The research provides a solid foundation for future works in fair ML. The insights gained from this study contributes to the ongoing efforts to develop more equitable and trustworthy AI systems.

**6. Author Contributions (Mandatory Section)**

**John Arthur Junior**: Conceptualization, methodology, software, implementation, experimental design, data curation and writing (draft preparation).

**Abdul Lafeet Yussif**: Methodology support, experimental validation, data preprocessing, software verification and writing (review and editing).

**Maame G. Asante-Mensah**: Supervision, conceptual validation, data preprocessing, software verification, methodology refinement and overall research oversight.

**Charles R. Haruna**: Formal analysis, results interpretation and writing (review and editing).

**Sandro Amofa**: Investigation, experimental execution, results visualisation and writing (review and editing).

**Elliot Attipoe**: Literature review, comparative analysis and results interpretation.

## 7. Funding

Not applicable

### 8. Ethical statement

Not applicable

## 9. Data availability statement

The data used in this study are publicly available benchmark datasets and were not generated by the authors. Specifically, the Adult Income dataset is available from the UCI Machine Learning Repository, the ProPublica COMPAS dataset is available from ProPublica's public data repository, and the Framingham Heart Study dataset is available from publicly accessible research data repositories commonly used in machine learning research. All datasets are provided in de-identified form and can be accessed without restriction for research purposes. No additional source data were generated or deposited as part of this study. Source data not available for this article. There are no legal, ethical, or privacy restrictions preventing access to the datasets, as all data used are anonymized and released under terms that permit academic research use.

## 10. Conflict of Interest

The authors declare no conflict of interest.

### 11. Acknowledgments

We would like to express our profound gratitude to the researchers in our department, for their invaluable guidance, constructive feedback, and continuous encouragement throughout the course of this study. We are also grateful to the faculty and staff of the Department of Computer Science and Information Technology, University of Cape Coast, for providing the academic environment and resources that made this work possible. Special thanks go to colleagues and peers for their insightful discussions and support during the development and evaluation of this study.

## 12. References

[1] D. Pessach and E. Shmueli, "Algorithmic fairness," in Machine Learning for Data Science Handbook: Data Mining and Knowledge Discovery Handbook, Springer, 2023, pp. 867–886.

[2] X. Ferrer, T. Van Nuenen, J. M. Such, M. Coté, and N. Criado, "Bias and discrimination in AI: a cross-disciplinary perspective," arXiv preprint arXiv:2008.07309, 2020.

[3] X. Wang, Y. Zhang, and R. Zhu, "A brief review on algorithmic fairness," Management System Engineering, vol. 1, no. 1, p. 7, 2022.

[4] J. Dressel and H. Farid, "The accuracy, fairness, and limits of predicting recidivism," Sci Adv, vol. 4, no. 1, p. eaao5580, 2018.

[5] D. Yeung, I. Khan, N. Kalra, and O. Osoba, Identifying systemic bias in the acquisition of machine learning decision aids for law enforcement applications. JSTOR, 2021.

[6] A. Khademi and V. Honavar, "Algorithmic bias in recidivism prediction: A causal perspective (student abstract)," in Proceedings of the AAAI Conference on Artificial Intelligence, 2020, pp. 13839–13840.

[7] H. Bubakr and C. Baber, “Using the Toulmin Model of Argumentation to Explore the Differences in Human and Automated Hiring Decisions.,” in VISIGRAPP (2: HUCAPP), 2020, pp. 211–216.

[8] S. Verma, M. Ernst, and R. Just, “Removing biased data to improve fairness and accuracy,” arXiv preprint arXiv:2102.03054, 2021.

[9] M. Wan, D. Zha, N. Liu, and N. Zou, “In-processing modeling techniques for machine learning fairness: A survey,” ACM Trans Knowl Discov Data, vol. 17, no. 3, pp. 1–27, 2023.

[10] S. Barocas, M. Hardt, and A. Narayanan, Fairness and machine learning: Limitations and opportunities. MIT press, 2023.

[11] G. Pleiss, M. Raghavan, F. Wu, J. Kleinberg, and K. Q. Weinberger, “On fairness and calibration,” Adv Neural Inf Process Syst, vol. 30, 2017.

[12] I. Žliobaitė and B. Custers, “Using sensitive personal data may be necessary for avoiding discrimination in data-driven decision models,” Artif Intell Law (Dordr), vol. 24, no. 2, pp. 183–201, 2016.

[13] S. Boulitsakis-Logothetis, “Fairness-aware naive Bayes classifier for data with multiple sensitive features,” arXiv preprint arXiv:2202.11499, 2022.

[14] Y. Choi, G. Farnadi, B. Babaki, and G. Van den Broeck, “Learning fair naive bayes classifiers by discovering and eliminating discrimination patterns,” in Proceedings of the AAAI Conference on Artificial Intelligence, 2020, pp. 10077–10084.

[15] T. Jang, P. Shi, and X. Wang, “Group-aware threshold adaptation for fair classification,” in Proceedings of the AAAI Conference on Artificial Intelligence, 2022, pp. 6988–6995.

[16] H. Zhang, “The optimality of naive Bayes,” Aa, vol. 1, no. 2, p. 3, 2004.

[17] T. Jang, P. Shi, and X. Wang, “Group-aware threshold adaptation for fair classification,” in Proceedings of the AAAI Conference on Artificial Intelligence, 2022, pp. 6988–6995.

[18] E. Ferrara, “Fairness and bias in artificial intelligence: A brief survey of sources, impacts, and mitigation strategies,” Sci, vol. 6, no. 1, p. 3, 2024.

[19] C. J. S. Barr, O. Erdelyi, P. D. Docherty, and R. C. Grace, “A Review of Fairness and A Practical Guide to Selecting Context-Appropriate Fairness Metrics in Machine Learning,” arXiv preprint arXiv:2411.06624, 2024.

[20] S. Friedler, C. Scheidegger, and S. Venkatasubramanian, “Certifying and removing disparate impact,” arXiv preprint arXiv:1412.3756, 2014.

[21] A. Agarwal, H. Agarwal, and N. Agarwal, “Fairness Score and process standardization: framework for fairness certification in artificial intelligence systems,” AI and Ethics, vol. 3, no. 1, pp. 267–279, 2023.

[22] R. Long, “Fairness in machine learning: Against false positive rate equality as a measure of fairness,” J Moral Philos, vol. 19, no. 1, pp. 49–78, 2021.

[23] E. Chzhen, C. Denis, M. Hebiri, L. Oneto, and M. Pontil, “Leveraging labeled and unlabeled data for consistent fair binary classification,” Adv Neural Inf Process Syst, vol. 32, 2019.

[24] Hardt, M., & Negri, A. (2009). Commonwealth. Harvard University Press.

[25] Corbett-Davies, S., Gaebler, J. D., Nilforoshan, H., Shroff, R., & Goel, S. (2023). The measure and mismeasure of fairness. Journal of Machine Learning Research, 24(312), 1-117.